\documentclass[letterpaper, 10 pt, conference]{IEEEtran}
\IEEEoverridecommandlockouts
\usepackage{balance}
\usepackage{cite}
\usepackage{amsmath,amssymb,amsfonts,bbm,mathtools}
\usepackage{hyperref}
\usepackage{bm}

\usepackage[ruled, linesnumbered]{algorithm2e}
\SetAlCapNameFnt{\footnotesize}
\SetAlCapFnt{\footnotesize}

\usepackage{graphicx,subcaption}
\usepackage{textcomp}
\usepackage{xcolor}
\usepackage{comment}
\usepackage{multirow}
\usepackage{enumitem}
\usepackage{lipsum}
\usepackage{arydshln}

\def\BibTeX{{\rm B\kern-.05em{\sc i\kern-.025em b}\kern-.08em
    T\kern-.1667em\lower.7ex\hbox{E}\kern-.125emX}}

\let\oldnl\nl
\newcommand{\nlnonumber}{\renewcommand{\nl}{\let\nl\oldnl}}

\newcommand{\svec}[1]{\bm{#1}}
\newcommand{\smat}[1]{\bm{#1}}
\newcommand{\Tr}{^{\top}}

\newcommand{\norm}[1]{\left\lVert #1 \right\rVert}
\newcommand{\ctrl}{\svec{x}}
\newcommand{\obs}{\svec{z}}

\begin{document}

\title{\LARGE \bf Live Thread Reconstruction for Autonomous Manipulation}

\author{
% Anonymous Authors
{Emma Huang$^{1}$,  Omar Hernandez$^{1}$, Zih-Yun Chiu$^{2}$, Michael C. Yip$^{1}$ \IEEEmembership{Senior Member, IEEE}}% <-this % stops a space
\thanks{$^{1}$Department of Electrical and Computer Engineering, University of California San Diego, La Jolla, CA 92093 USA. {\tt\small\{emh002, o2hernandez, yip\}@ucsd.edu}}% <-this % stops a space
\thanks{$^{2}$Department of Computer Science, Johns Hopkins University, Baltimore, MD 21218 USA. {\tt\small zchiu@jhu.edu}}% <-this % stops a space
}

\maketitle
\begin{abstract}
Autonomous robotic surgery has drawn increased attention in recent years due to its potential to relieve surgeons of time-consuming, mentally straining, and tedious tasks, particularly during prolonged operations.
One suitable task for automation is suturing, which involves tracking the thread as it deforms during needle retrieval, de-tangling, knot tightening, and other tasks. Unlike rigid-object manipulation, deformable thread manipulation has no fixed pose to track and effectively infinite degrees of freedom, making its configuration difficult to predict and requiring the thread's state to be re-estimated rather than assumed. At the same time, thread manipulations involve sequential motions that depend on past and current thread contact points, which must be accurately and continuously tracked. This paper presents an image-to-manipulation pipeline that concurrently generates live thread reconstruction while conducting safe and optimized sequential manipulations to achieve long-horizon tasks.

\end{abstract}

\section{Introduction}
Autonomous robotic surgery has drawn increased attention in recent years due to its potential to relieve surgeons of time-consuming, tedious tasks, particularly during prolonged operations. By improving the speed and quality of surgeries, automation can help treat more patients and mitigate the impact of the global surgeon shortage. One highly suitable task for automation is suturing, which requires complex manipulation of the highly deformable thread during retrieval, pulling, knot tightening, de-tangling, and loop formation. Deformable object manipulation is itself a longstanding open problem in robotics: unlike rigid objects, a surgical thread has effectively infinite degrees of freedom, so its state cannot be summarized by a fixed pose and must instead be continuously re-estimated as it bends, crosses itself, and slides through the gripper. Executing suturing maneuvers successfully therefore requires the thread to be continuously tracked and accurately handled despite highly unpredictable configurations and noisy sensing.

Despite recent advancements, current methods remain limited in their scope and applicability. Many approaches initialize manipulation with the thread or needle already secured in the gripper, entirely bypassing the difficult but critical step of deformable object handling. Other techniques rely on static thread reconstruction~\cite{huang2026reel, joglekar2025autonomous}, which restricts the system to simple manipulations and requires predictable thread configurations to succeed, since re-solving the full configuration space of a deformable thread from scratch at each step is intractable for long-horizon tasks. Furthermore, while imitation learning has been applied to needle pickup and knot tying~\cite{kim2024srt, Kim2025imitationsurgery}, these models largely operate as black boxes, offering little control over essential safety constraints required in the operating room.

To bridge this gap in surgical robot automation, we propose an integrated pipeline that advances from raw images to autonomous manipulation by live tracking the thread through long horizon tasks. Our approach consists of four primary components: (1) initial reconstruction of the thread, needle, and surgical environment; (2) autonomous, continuous reconstruction utilizing a Kalman Filter for noise reduction and temporal consistency; (3) constrained thread grasp-point optimization based on safety and task criteria; and (4) autonomous thread manipulation to complete the designated tasks. % Reliable and fast thread reconstruction is extremely valuable, as it assists surgeons in locating obscure suture threads, determining future grasping goals, and automating tedious manipulation tasks, thereby preserving the surgeon's time and energy for more critical surgical phases.

\begin{figure}[t!]
    \centering
    \includegraphics[width=0.99\linewidth]{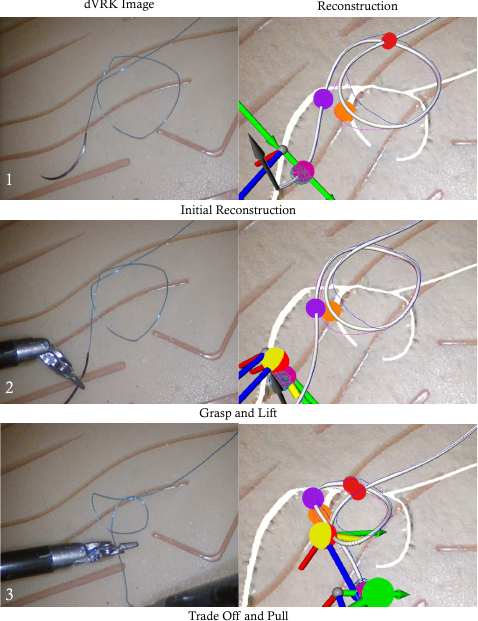}
    \caption{The thread reconstruction remains consistent through common surgical thread manipulations such as knot tightening while handling occlusions from the tool and self intersections causing ambiguity.}
    \label{fig:title-image}
\end{figure}

Our main contributions toward advancing medical robotics are as follows:
\begin{itemize}
    \item Real-time thread reconstruction that remains robust to complex intersections and occlusions.
    \item The integration of thread-gripper interaction modeling to improve both reconstruction reliability and overall manipulation efficiency.
    \item Experimental evaluation of the proposed thread manipulation methods on needle pick-up and knot tightening, demonstrating reliable performance under continuous, live reconstruction.
\end{itemize}

\section{Methods}
\begin{figure*}[t!]
    \centering
    \includegraphics[width=0.99\linewidth]{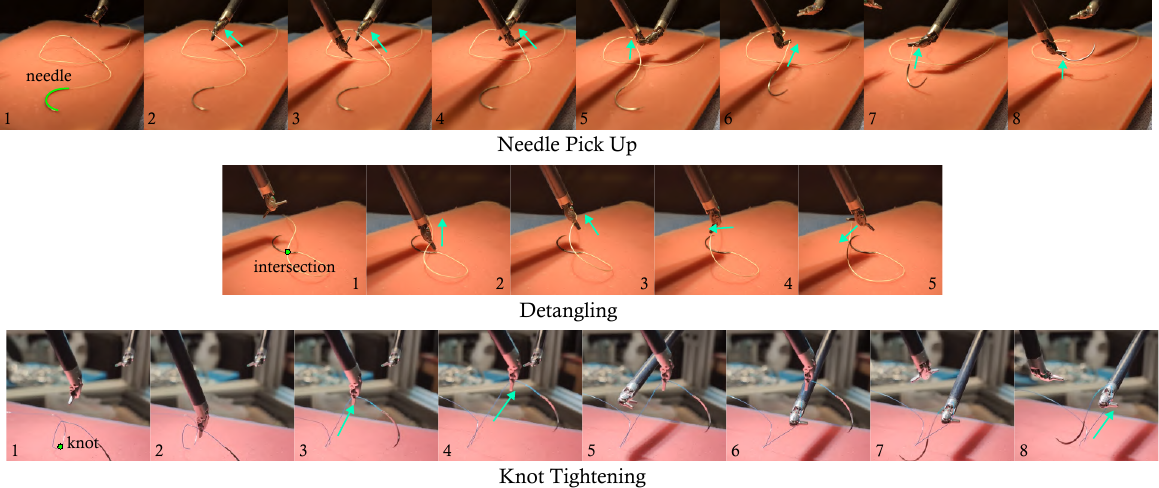}
    \caption{The thread reconstruction is showcased on three tasks commonly seen in surgery including needle pick up, thread de-tangling, and knot tightening. For each task, the key object such as the needle, intersection, and knot shape the formation of manipulation motions.}
    \label{fig:manipulation-showcase}
\end{figure*}

\subsection{Thread Reconstruction Overview}
For each frame $k$, a rectified stereo pair and a binary segmentation mask $\mathcal{M}$ drive the reconstruction pipeline: (1) Keypoint Selection, (2) Intersection Detection, (3) Keypoint Ordering, (4) Spline EKF, and (5) Spline Initialization. Frame $1$ uses graph-based ordering and quadratic programming (QP)~\cite{joglekar2025autonomous}, while subsequent frames utilize the EKF posterior as a reference.
\subsection{Keypoint Selection}\label{subsec:select}
Following~\cite{joglekar2025autonomous}, stereo matching scores disparities by sum of squared differences, with disparity taken as the center of the contiguous plateau, $d(\svec{p})=\tfrac{1}{2}(d_\ell+d_h)$, to avoid bias on epipolar-parallel segments. Disparity confidence $\rho(\svec{p})$ comes from the next-best score margin and plateau width; disparities are converted to 3-D, gated by a workspace band filter, and clustered~\cite{joglekar2025autonomous}. Clusters lacking reliable depth interpolate their $Z$ coordinate from the two nearest reliable anchor clusters, weighted by inverse Dijkstra distance over the adjacency graph.

\subsection{Intersection Detection}\label{subsec:intersection}
A 2-D structure tensor $\mathbf{T}$ evaluates local keypoint neighborhoods $\mathcal{N}$, classifying crossing candidates by principal eigenvalue ratio $\lambda_1/\lambda_0 > 0.2$. RANSAC extracts dense pixel arms $\svec{a}_{j,m}$ with centroid $\svec{\mu}$ and direction $\svec{\nu}$; to filter hairpin bends and parallel threads, valid arms must spatially straddle the crossing centroid on both sides of $\svec{\mu}$ along $\svec{\nu}_{j,m}$. The crossing center is the intersection of the two verified arms' axes.

\subsection{Keypoint Ordering}\label{subsec:order}
\emph{Cold ordering} assigns parameters $s_i$ via shortest-path traversal on the adjacency graph~\cite{joglekar2025autonomous}. \emph{Warm ordering} instead matches keypoints to the previous posterior, gated on lateral distance, depth distance, Mahalanobis fit, ambiguity against alternative matches, and proximity to crossing centers; these combine into a match quality $q_i \in [0,1]$. Keypoints near crossing centers are stripped of initial matches and later interpolated along their nearest arm's axis if they pass a 3-D $\chi^2$ test.

\subsection{Initialization and Spline-State Extended Kalman Filter}\label{subsec:ekf}
For the initial frame, we instead solve a QP over the B-spline control points $\svec{y}$~\cite{joglekar2025autonomous}, minimizing bending energy and temporal deviation subject to keypoint-derived bounding constraints:
\begin{gather}
\min_{\svec{y}} \svec{y}\Tr\smat{D}\Tr\smat{W}\smat{D}\svec{y} + \lambda\norm{\smat{E}(\svec{y}-\svec{y}^{\mathrm{prior}})}^2 \\
\text{s.t.} \quad \svec{\ell}_i \le \smat{B}(s_i)\svec{y} \le \svec{u}_i \nonumber
\end{gather}

The state is the control polygon $\ctrl$ of a cubic spline, giving a linear observation model $\obs_i=\smat{H}_i\ctrl+\svec{\eta}_i$. The process model warps control points based on tool motion, exponentially damped by distance to the tool tip, with process noise scaled by motion magnitude $m_k$. The update uses standard information form, inflating measurement precision by match quality and confidence, $\smat{R}_i^{-1} \leftarrow \max(c_i q_i, w_{\min}) \smat{R}^{-1}$, and incorporates smoothness, curvature, arc-length, and knot-count shape priors as pseudo-measurements; interior control points with low observational support $\zeta_j = \sum_{i} |b_j(s_i)|$ are frozen, and unsupported ends are trimmed.
Bounds $[\svec{\ell}_i, \svec{u}_i]$ are relaxed proportionally to $1-q_i$ so poorly matched keypoints cannot distort the fit.

\subsection{Manipulation Methods}
\subsubsection{Needle Pick-Up}
The objective is to walk the grip up the thread until the needle at the free tip can be reached, selecting the grasp point nearest the needle by bypassing crossing predicates. Once both arms grip the thread, the arm closest to the needle becomes the static anchor and the other the active holder; the holder pulls the thread through the anchor with a stroke lifting orthogonally to the anchor's shaft, and rounds of lifting, pulling, pinning, releasing, re-centering, and regrasping repeat until the tip margin is reached.

\subsubsection{De-tangling}
De-tangling opens self-crossings by identifying and manipulating the highest-standing strand, evaluating thread-sample height along an arc band interior to the loop and targeting the crossing with the greatest upper-pass height, with grasp candidates restricted to the loop interior and a defined arc band. The arm grasps and lifts the strand, translates over the crossing, and rolls about the shaft axis to untwist the loop, then withdraws along the loop's tangent and transitions into the pull-through rounds used in needle pick-up until all crossings are resolved.

\subsubsection{Knot Tightening}
Knot tightening removes slack between an existing hold and the knot, restricting grasp candidates to the span between hold and knot while rejecting loop-interior points and enforcing keep-out spheres around the knot and crossings. The pull stroke follows the thread's tangent on the needle side of the knot, skewed for horizontal progress and tilted upward to lift the knot during cinching; alternating trade rounds—receiver grasps and closes before giver releases, so tension is never fully dropped—loop until the hold walks up to the knot. This method is shown in Figure~\ref{fig:title-image}.

\section{Experiments and Results}
We evaluate our thread and needle manipulation pipeline on two tasks, shown in Figure~\ref{fig:manipulation-showcase}: needle pick up and suture knot tightening. The needle pick up success rate is compared with static (single-frame)~\cite{huang2026reel} thread reconstruction methods, across easy/medium and hard configurations; knot tightening is evaluated under live reconstruction only as no previous benchmark exists. Results are shown in Table~\ref{tab:main-tasks}.

\begin{table}[h]
\centering
\caption{Manipulation Task Success Rates}
\begin{tabular}{l | l r : l r}
 & \multicolumn{2}{c}{Static} & \multicolumn{2}{c}{Live} \\
\hline
Setup & Easy+Medium & Hard & Easy+Medium & Hard \\
\hline
Needle Pick Up & 18/20 & 7/10 & 13/15 & 14/15 \\
Tightening & - & - & 15/15 & 13/15 \\
\end{tabular}
\label{tab:main-tasks}
\end{table}

Needle pick up succeeds on 18/20 (90\%) easy/medium trials under static reconstruction and 13/15 (87\%) under live reconstruction, comparable performance for the less ambiguous configurations. The two diverge on hard configurations: live reconstruction reaches 14/15 (93\%), while static reconstruction drops to 7/10 (70\%). We attribute this gap to live reconstruction's ability to persist through gripper occlusion and re-localize the thread mid-grasp, conditions under which a single-frame static reconstruction cannot recover. Knot tightening under live reconstruction succeeds on 15/15 (100\%) easy/medium trials and 13/15 (87\%) hard trials.

Qualitatively, once a reconstruction is initialized on an unambiguous frame, we observe it remains stable when the grippers occlude large segments of the thread, the thread is moved out of frame, or the disparity becomes noisy. We plan to quantify this stability, and the reconstruction's accuracy relative to Joglekar et al.~\cite{joglekar2025autonomous}, in ongoing work.

Future work includes:
\begin{itemize}
    \item quantitative reconstruction accuracy against Joglekar et al.~\cite{joglekar2025autonomous}, comparing projected reconstruction to segmentation masks across visibility conditions (fully visible, minimally occluded) and recovery scenarios (out-of-frame for static, tool occlusion for live)
    \item quantitative results on additional manipulation tasks including thread de-tangling.
\end{itemize}

\section{Conclusion and Future Work}
% roughly what is expected
In this paper, we provide a thread reconstruction and manipulation pipeline that sustains accurate thread grasping under continuous (live) manipulation. The pipeline is tested on needle pick up and suture knot tightening, where live reconstruction improves hard-configuration needle pick up success to 14/15 (93\%) over 7/10 (70\%) with static reconstruction. Qualitatively, once initialized, our reconstruction remains stable under gripper occlusion, out-of-frame thread motion, and noisy disparity — improvements over prior thread reconstruction methods that we plan to quantify directly in ongoing work. The main drawback include its inability to provide accurate reconstructions on epipolar-parallel segments of the thread, which limit the range of manipulations possible. Regardless of this limitation, this method showcases potential for all sorts of thread manipulations with the ability to specify specific criteria and constraints that will be essential for real surgical deployment.
% \section*{Acknowledgment}
% leave empty

%\clearpage
%\balance
\bibliographystyle{IEEEtran}
\bibliography{ref}

\end{document}